\documentclass[a4paper,conference]{IEEEtran}
\usepackage{cite}
\ifCLASSINFOpdf
   \usepackage[pdftex]{graphicx}
   \DeclareGraphicsExtensions{.pdf,.jpeg,.png}
\else
   \usepackage[dvips]{graphicx}
   \DeclareGraphicsExtensions{.eps}
\fi
\usepackage{amsmath}
\ifCLASSOPTIONcompsoc
  \usepackage[caption=false,font=normalsize,labelfont=sf,textfont=sf]{subfig}
\else
  \usepackage[caption=false,font=footnotesize]{subfig}
\fi
\usepackage{fixltx2e}
\begin{document}
\title{Convolutional STN for Weakly Supervised Object Localization}

%\author{\IEEEauthorblockN{Akhil Meethal}\\
%\IEEEauthorblockA{Laboratory of Imaging, Vision and Artificial Intelligence (LIVIA)\\
%École de technologie superieure\\
%Universite du Québec\\
%Email: akhilpm135@gmail.com}\\
%\and
%\IEEEauthorblockN{Marco Pedersoli}\\
%\IEEEauthorblockA{Laboratory of Imaging, Vision and Artificial Intelligence (LIVIA)\\
%École de technologie superieure\\
%Universite du Québec\\
%Email: marco.pedersoli@etsmtl.ca}\\
%\and
%\IEEEauthorblockN{Soufiane Belharbi}
%\IEEEauthorblockA{Starfleet Academy\\
%San Francisco, California 96678--2391\\
%Email: soufiane.belharbi.1@ens.etsmtl.ca}}


\author{\IEEEauthorblockN{Akhil Meethal, Marco Pedersoli, Soufiane Belharbi and Eric Granger}\\
\IEEEauthorblockA{Laboratory of Imaging, Vision and Artificial Intelligence (LIVIA)\\
Dept. of Systems Engineering\\
École de Technologie Superieure, Université du Québec\\
Montreal, Canada\\
Email: akhilpm135@gmail.com, marco.pedersoli@etsmtl.ca, soufiane.belharbi.1@ens.etsmtl.ca, eric.granger@etsmtl.ca}}


\maketitle

% !TEX root=main.tex

\begin{abstract}
Weakly-supervised object localization is a challenging task in which the object of interest should be localized while learning its appearance. State-of-the-art methods recycle the architecture of a standard CNN by using the activation maps of the last layer for localizing the object. While this approach is simple and works relatively well, object localization relies on different features than classification, thus, a specialized  localization mechanism is required during training to improve performance. In this paper, we propose a convolutional, multi-scale spatial localization network that provides accurate localization for the object of interest. Experimental results on CUB-200-2011 and ImageNet datasets show that our proposed approach provides competitive performance for weakly supervised localization. 
%\sbx{Max pages: 9 excluding biblio. the rest needs to be in supp.mat.}
\end{abstract}


% \fi
\IEEEpeerreviewmaketitle

% !TEX root=main.tex

\section{Introduction}
\label{sec:introduction}

% general introduction to wsol.
Object localization is a key task in many computer vision applications such as autonomous driving~\cite{aut_driving-Chen-2015}, pedestrian detection~\cite{ped_detection-Wang-2018}, and earth vision \cite{earth_vision-Xia-2018}. %These applications generally employ fully supervised object detectors. 
In a standard learning setup, object localization requires full supervision, i.e, the class label and bounding box annotation for each object instance present in the image \cite{rfcn-Dai-2016,fast_rcnn-Girshick-2015,rcnn-Girshick-2016,fpn-Lin-2017,ssd-Liu-2016,faster_rcnn-Ren-2015,yolo-Redmon-2016,yolo_9000-Redmon-2017}. However, obtaining bounding box annotations is time consuming, in particular for large real-world image datasets. Moreover, human annotation can be subjective. 
% needs more expanding for weakly supervised learning.
%Recently, weakly supervised learning (WSL) paradigm has emerged in machine learning to reduce the need to fine-grained annotation \cite{zhou2017brief}. 
%
Recently, several weakly supervised learning (WSL) techniques have been proposed for object localization, to alleviate the need for such expensive fine-grained annotations \cite{ spg-Zhang-2018, acol-Zhang-2018, hide_and_seek-Singh-2017}. These techniques are applied in scenarios where supervision is either incomplete, inexact or ambiguous~\cite{zhou2017brief}. 
%In weakly supervised object localization (WSOL), the inexact supervision comes often with the global image level label only allowing to surpass the need to bounding boxes. Such weak supervision allows building good performing WSOL techniques and gain large popularity in computer vision community \cite{cam, acol, hide_and_seek, adl, spg, wsddn, wccn, wsl_mil, oicr}.
%\cite{wsddn,wsl_mil,wccn,hide_and_seek}.
% techniques have been proposed for object localization, to alleviate the need for such expensive fine-grained annotations \cite{wsddn,wsl_mil,wccn,hide_and_seek}. 
% They are applied in scenarios involving either (1) incomplete supervision (when only a small subset of training data has labels, although unlabeled data is abundant), (2) inexact supervision (when training with labeled data with coarse labels), and (3) ambiguous or inaccurate supervision (when labels may suﬀer from errors or noise) \cite{zhou2017brief}. 
The inexact supervision scenario is often considered for object localization tasks, where training datasets only require global image-level annotations, i.e, the class label for each object in an image. 
%In object localization task, WSL comes often with the global image label supervision, i.e., the class of the object.
%Under this scenario, powerful techniques for multiple-instance learning (MIL) (Carbonneau et al., 2018; Cheplygina et al., 2019; Wang et al., 2018; Zhou, 2004) are generally considered, where individual instance labels (e.g. image pixels, segments or patches) are not observable or do not belong to well-deﬁned classes – training instances are grouped into sets (e.g. images), and supervision is only provided for sets of instances.
%Under this scenario, WSOL techniques have gained much popularity in the computer vision community.
For this reason, weakly supervised object localization (WSOL) techniques based on image-level annotations have gained much popularity in the computer vision community 
\cite{wsddn-Bilen-2016,wsl_mil-Cinbis-2016,wccn-Diba-2017,oicr-Tang-2017,cam-Zhou-2016}. 

% \begin{figure}
% \center
% \begin{tabular}{cc}
% \bmvaHangBox{\fbox{\includegraphics[width=0.4\linewidth]{images/stdconvvscstn.pdf}}}&
% \bmvaHangBox{\fbox{\includegraphics[height=6cm, width=0.4\linewidth]{images/transform_adapt.png}}}\\
% (a)&(b)
% \end{tabular}
% \caption{(a) An illustration of the difference between standard convolution and CSTN. $P$ and $P^{\prime}$ are the depth of the feature maps. (b) Samples of bounding boxes localized using our proposed Convolutional STN. The images illustrate how our transform can adapt the receptive field box to improve localization. The receptive field box is shown in blue, transformed box in red, and the ground-truth in green.}
% \label{conv_stn_action}
% \end{figure}

\begin{figure}[tb]
\centering
\includegraphics[scale=0.19]{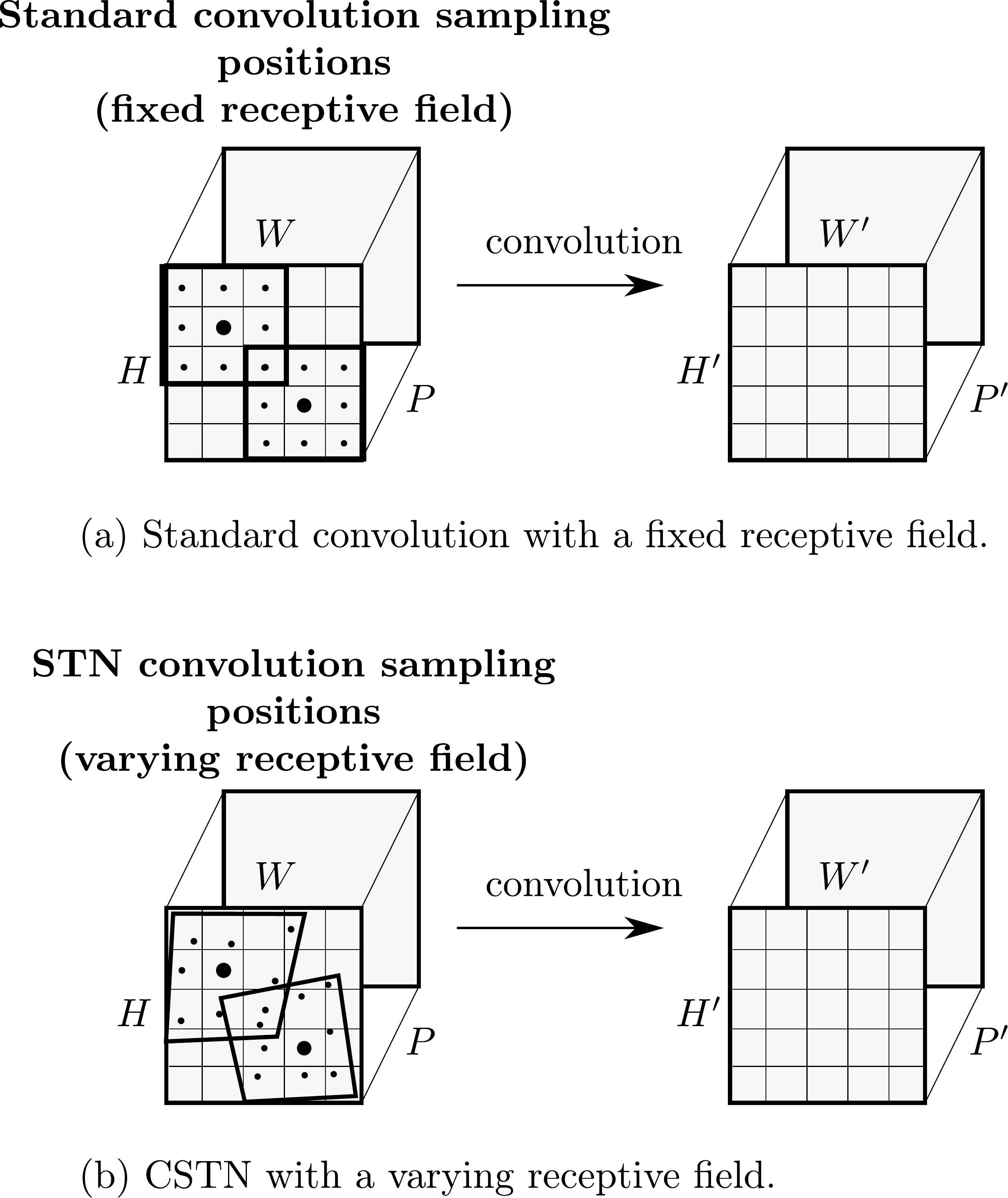}
\caption{ An illustration of the difference between the standard convolution and Convolutional STN. $P$ and $P^{\prime}$ are the depth of the feature maps}
\label{conv_stn_action}
\end{figure}

%%% CHALLENGES WITH SOA
%%%%%%%%%%%%%%%%%%%%%%%%
% convolution issue of adaptation to variations, and how it is dealt with. related work.
Deep convolutional neural networks (CNNs) with Class Activation Maps (CAM) \cite{cam-Zhou-2016} are a prominent solution in the literature for WSOL problems~\cite{hide_and_seek-Singh-2017,acol-Zhang-2018,cam-Zhou-2016}.  They use spatial class-specific localization maps where high activations indicate the location of the corresponding object of the class. %Most recently proposed WSOL techniques attempt to improve the CAM to address its limitations~\cite{ hide_and_seek-Singh-2017, acol-Zhang-2018}.  
CAMs are obtained through standard convolution, and as such, are limited in their ability to accommodate large and unknown transformations, and variations in object scale, orientation, and pose. %\cite{deformable_conv-Dai-2017,hinton2011Transforming,stn-Jaderberg2015,Kosiorek2019stacked,sabour2017dynamic,zhao2015stacked}. 
% This paragraph causes a lot of problems.
% These commonly-known limitations are mainly due to the fixed structure of the convolution operation. In practice, this issue can be alleviated using (1) data augmentation, and (2) transformation-invariant operations. In the first case, large  training datasets  are generated  through the  application of known and predefined affine transformations. Exposing the model to such data is expected to allow it to learn these diverse variations. While this approach is practical, data augmentation still requires to manually set fixed and known parametric transformations. This prevents generalization to unknown but potentially beneficial transformations for different visual recognition tasks. In the second case, the CNN performs transformation-invariant operations such as deformable convolution operation \cite{deformable_conv-Dai-2017} and max-pooling \cite{boureau2010theoretical}. While such operations allow the model to intrinsically handle object deformations, and to provide robustness for small translations, these operations remain predefined operations. Moreover, some operations (like max-pooling) are mechanical, and do not adapt to the task in hand, which reduces their flexibility. Moreover, designing each transformation separately could be difficult and overwhelming in term of computation and system complexity.
%%% PROPOSED SOLUTION: learnable transformation-invariant convolution operation.
%%%%%%%%%%%%%%%%%%%%%%
Learning a transformation-invariant operation that can simultaneously handle different transformations is desirable for visual recognition systems. Spatial Transformer Networks (STNs) \cite{stn-Jaderberg2015} have been proposed recently as a differentiable module that allows for spatial transformation of data within a CNN without manual intervention. This provides the network with flexibility in term of adaption to the input image variations. Since the location of the activation in CAMs are intrinsically dependent to the convolution operation, \emph{flexible} convolution operation that \emph{adapts} to scale, orientation, and other possible variations are preferable. In this work, we investigate the use of STNs \cite{stn-Jaderberg2015} as an \emph{adaptive convolution operation} to replace the standard convolution. We refer to this operation as Convolutional STN (CSTN). This adaptation is achieved through the application of an STN convolution over each location. STN model learns affine transformations that can cover different variations including translation, scale, and rotation, allowing to better attend different object variations. This provides additional flexibility compared to standard convolution. Figure~\ref{conv_stn_action} illustrates the difference between both types of convolution. In standard convolution, the sampling grid of the convolution is fixed (hence, it has fixed receptive field), while our CSTN transforms the sampling grid using spatial transformers and samples the input feature map from the resulting locations, allowing it to have a varying receptive field.

%CAM can be essentially obtained from a classification convolutional network as a spatial map where high activations indicate the location of the corresponding object of the class. 
%CAMs are obtained though performing standard spatial convolution operation with fixed receptive field and orientation over the spatial representation of the previous layer. Such fixed operation my lead to missing objects with different scales and/or orientations which are two important factors for accurate object localization. 

% this argument needs more work.
%\focus{Although a CAM network attempts to learn discriminative features to distinguish between the classes, different features are required for localization. We believe that specific components for learning the localization representation should be used in the WSOL for optimal performance. [needs work/references. i am not sure about this concept that classification/localization require different representation. there are many work \cite{oquab2015object,cam} that show that using only a classification supervision --which is what we do here, allow to localize objects, which is the main motivation to go to wsol. so the above sentence seems in contradiction with the goal of this paper/wsol in general.]} 

\begin{figure}[t]
 \centering
 \includegraphics[height=5.3cm, width=0.8\linewidth]{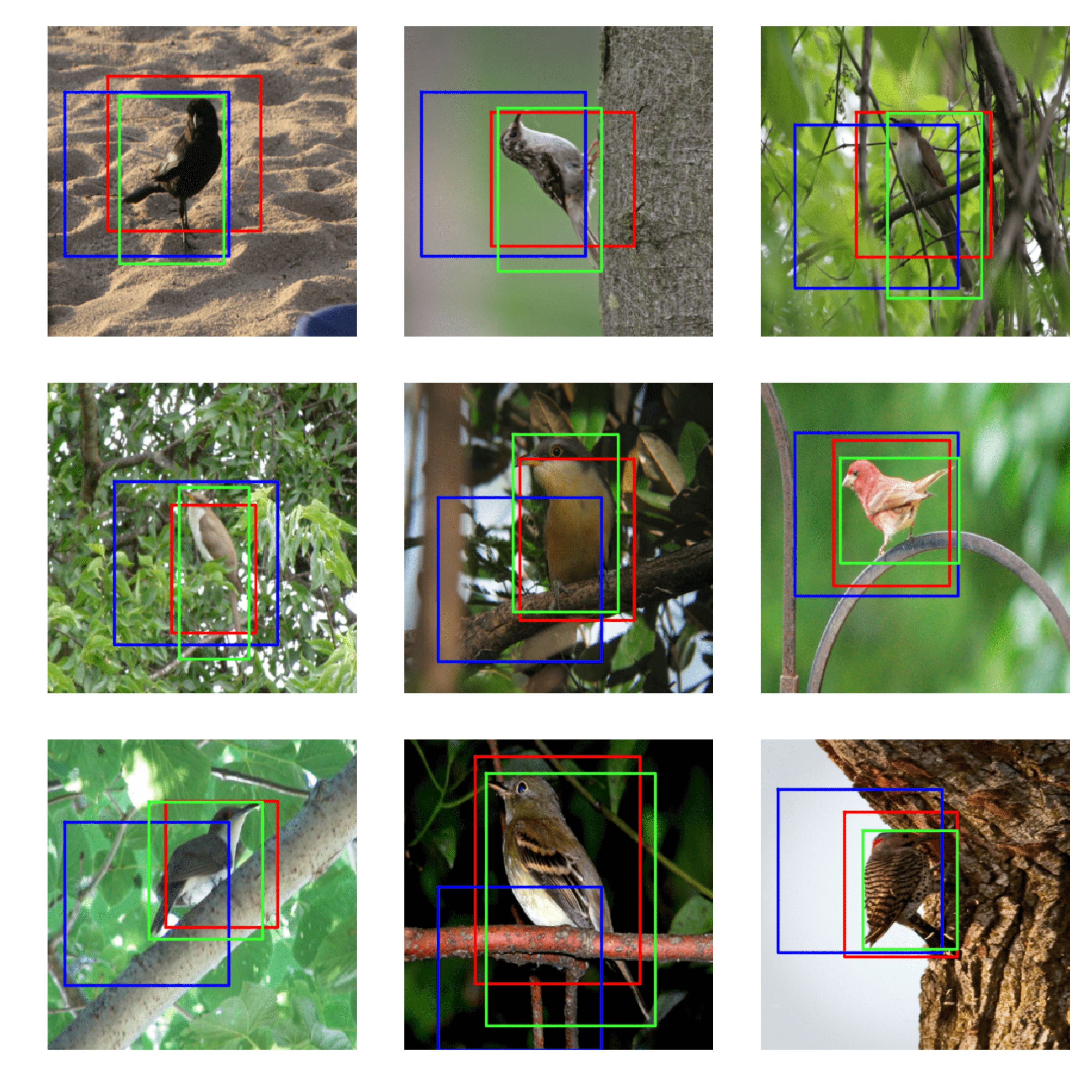}
\caption{Samples of bounding boxes localized using our proposed Convolutional STN. The images illustrate how our transform can adapt the receptive field box to improve localization. The receptive field box is shown in blue, transformed box in red, and the ground-truth in green.}
\label{transform_cover}
\end{figure}

While the CSTN is able to adapt to relatively small local variations, it still faces the issue of adapting to large variations in term of the receptive field. To alleviate this issue, we consider localizing objects of different scales at different levels (i.e, layers), using the Feature Pyramid Networks (FPN) \cite{fpn-Lin-2017}. The CSTN is applied at different levels of the feature pyramid. As the receptive field from the low layers can process only small regions of big objects, local convolution at that layer tend to localize small discriminative regions while missing the entire object. However, such layers are more adequate to localize small objects while high layers can miss them due to their large receptive field. To deal with this, an additional regularization term is introduced to drive  specific layers to compete for the right scale.  A joint probability over scale, location, and class is formulated based on the class scores through an aggregation process. We evaluate our approach on popular weakly supervised benchmark datasets and observe competitive performance on the localization task. Figure \ref{transform_cover} illustrates how the CSTN is able to adapt to improve the localization. To summarize, our main contributions are:
(1) a novel approach for WSOL with convolutional spatial transforms that explicitly learns to localize during classification;
(2) an adaptation of the FPN model \cite{fpn-Lin-2017} to weakly supervised settings for localizing objects of different scale, where the STN need to learn a small transform for the right scale;
(3) regularization techniques to prevent the localization from selecting discriminative object regions.

% Though CSTN was able to adapt the sampling grid to improve the localization, it was not able to generate precise transforms when the object size is far from the receptive field size. This can be due to the fact that the gradient based updates cannot adapt to a wide range of transforms according to the object scale, as it takes small steps while learning. So we proposed to localize objects of different scale from different levels of the convnet. To this end, we used Feature Pyramid Networks(FPN)\cite{fpn} in our weakly supervised settings. The conv STN is applied at different levels of the feature pyramid and the output is  scored with the same $1 \times 1$  classification head for all levels. As the receptive field from the lower levels can see the discriminative regions for bigger objects,  the classifier is more biased towards localizing the bigger objects from lower layers. An additional regularization term is introduced to avoid this bias by making the levels to compete for the right scale.  A joint probability over scale, location and class is generated from the class scores by concatenating them and applying a softmax. Empirical validation of the proposed approach is performed with CUB-200-2011 bird dataset \cite{cub_dataset} and ILSVRC 2012 \cite{imagenet} localization dataset.

\begin{figure*}[!t]
  \centering
  \includegraphics[scale=0.25]{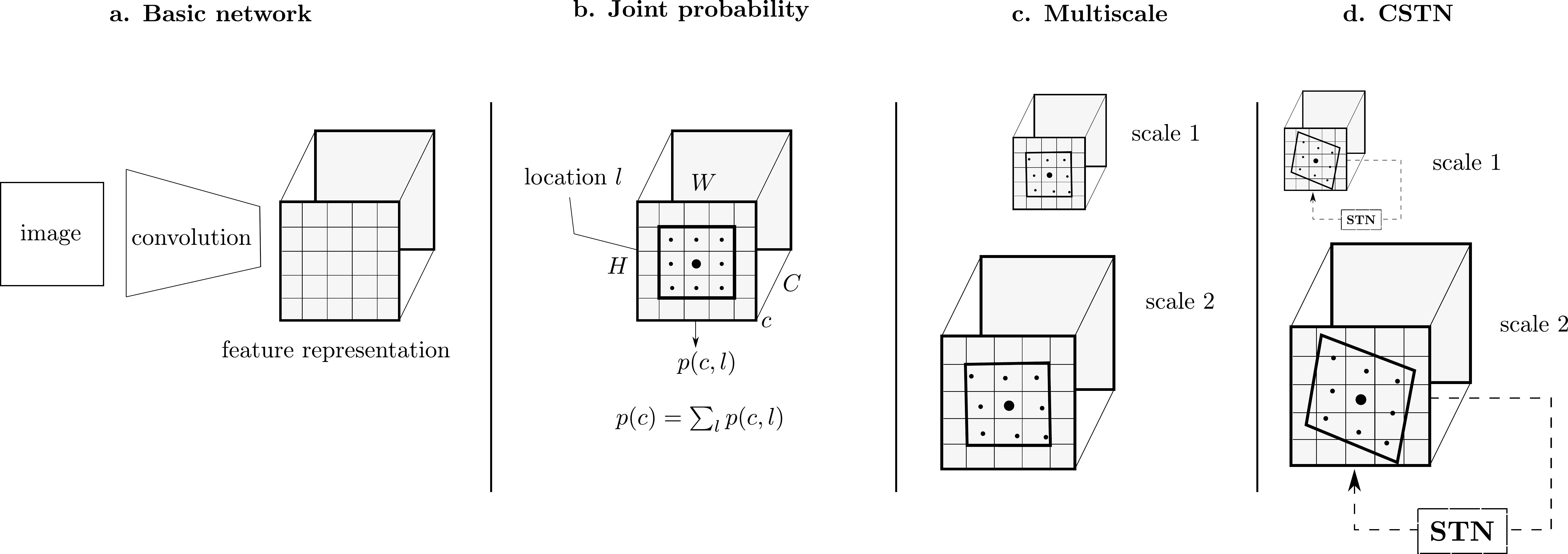}  % height=4cm, width=0.8\linewidth
  \caption{\textbf{Basic components of our system.} (a) One of the last convolutional layers of a CNN can already provide some information about the center of the object. (b) Our joint probability in location and classes is used to learn localization in a Weakly supervised manner (see text). (c) Using a multi-scale approach we can find not only the position of the object but also the scale (d) Adding our CSTN, we obtain a more refined localization of the object of interest.}
  \label{fig:fig-0-convolution}
\end{figure*}

% !TEX root=main.tex

\section{Related work}
\label{sec:related-work}

%- Topics: papers on WSOL, STN, multi-scale and FPN.
%- Deformable convolution

%- end with key challenges you pan to address

%\focus{a little redundant with the introduction. similar message and exact phrases.}

%Localizing objects in a fully supervised setting is studied extensively in the vision research, ranging from fast single shot detectors \cite{yolo, yolo_9000, ssd} to accurate two stage detectors \cite{rcnn, fast_rcnn, faster_rcnn, rfcn, fpn}. However, the need of expensive bounding box annotations is a fundamental bottleneck for designing practical object detectors through fully supervised learning. Consequently, weakly supervised methods that need only global image-level labels for object detection \cite{wsddn, w2f, wsl_mil, oicr, wccn} and localization \cite{cam, acol, spg, adl, hide_and_seek} have recently been gaining popularity. The dominant paradigm for weakly supervised object detection (WSOD) is Multiple Instance Learning (MIL). Whereas weakly supervised object localization methods rely on the class activation maps technique.  

\noindent \textbf{Weakly supervised object localization.} The Class Activation Map (CAM) is a pioneering technique in WSOL that was proposed by Zhou et al. \cite{cam-Zhou-2016}. It uses a simple and straightforward method to locate the strongly activated region using  a  fully convolutional classification network. Since it is primarily focused on achieving a high level of classification accuracy, its localization tends to correspond with the most discriminative object region. Most of the recent WSOL techniques propose updated versions of the CAM that can avoid the bias towards the discriminative region \cite{hide_and_seek-Singh-2017,acol-Zhang-2018,spg-Zhang-2018}. They typically seek to erase or hide the most discriminative region during training so that the classifier will focus on other relevant object regions. To achieve this, they leverage different strategies, like using multiple classifiers to localize complementary regions(ACoL)~\cite{acol-Zhang-2018}, self-produced guidance(SPG)~\cite{spg-Zhang-2018}, randomly hiding patches from the input image(HaS)~\cite{hide_and_seek-Singh-2017}. % and masking the discriminative region with dropout~\cite{adl-Choe-2019}.

\noindent \textbf{Deformable convolution.} The CSTN is in principle similar to deformable convolution proposed in  \cite{deformable_conv-Dai-2017}. To break the fixed geometry of a standard convolution, it learns a set of offsets for each position in the regular sampling grid. Deformable convolution has demonstrated improvements in object localization for the fully supervised object detectors \cite{deformable_conv-Dai-2017, deformable_conv_new-Zhang-2019}. However the deformation learned in this way is not a centralized one as each location in the sampling grid can move independently, resulting in irregular shape for the convolution. Thus it cannot be directly utilized for localization. Active convolution unit proposed in \cite{active_conv-Jeon-2017} attempts to learn the shape of the convolution. All these deformable convolution methods are studied in the fully supervised algorithms, we are the first to study it in a weakly supervised settings.

\noindent \textbf{Spatial transformers} The Spatial Transformer Networks (STN) has been proposed by Jaderberg et al.~\cite{stn-Jaderberg2015} to learn a global affine transform of its input feature map. Driven by the classification objective, this global transform allows locating relevant object regions. This can be interpreted as a soft attention mechanism. Since it is a generic learnable module that can transform its input with respect to the network objective, it has found many application in, e.g, image captioning~\cite{stn_captioning-Johnson-2018}, disentangled representation learning~\cite{stn_dientangled_rep-Detlefsen-2019} and image  composting~\cite{stn_image_composting-Lin-2018}. Our proposed method is applying it in a convolutional way to address the weakly supervised localization problem. As in~\cite{stn-Jaderberg2015}, the network objective of our CSTN is also to obtain accurate classification. However, the STN localization of objects on natural images (with a global transform) cannot easily cope with large object variations. As described in the next section, by learning local transforms from the appropriate level, our CSTN is able to provide better localization than the global transform.

%Invariance is often desirable in many computer vision tasks. SIFT, SURF etc, are some popular methods in the traditional vision literature for extracting invariant features. Spatial transformer networks\cite{stn} introduced a differentiable module for learning invariant features. 

% !TEX root=main.tex
\section{Overall Architecture}
\label{sec:proposal}
%One of the most popular approaches for weakly supervised object localization is CAM \cite{}. Weakly Supervised Object localization is performed as thresholding of the feature map generated by the last layer of a CNN. Even though very simple, it has several drawbacks. The main one is that the concept of object is lost 
To explain our architecture for WSOL, we start from the last convolutional layer of a CNN and show how it is used for object localization (see Fig.~\ref{fig:fig-0-convolution}(a)). Similar to one-stage object detection methods (e.g: SSD \cite{ssd-Liu-2016}, YOLO \cite{yolo_9000-Redmon-2017}), we consider the location of a filter as the rough center of the object. In one-stage detectors, this location is then associated with a set of class probabilities that defines which object is more likely to appear at that location and the coordinates of the object's bounding box, estimated as a regression. In our case, we do not have information about the bounding box of the object as our problem is weakly supervised (we only have image-level label). Thus, to go from object labels to image labels we need an aggregation mechanism as detailed in the next subsection.

\subsection{Joint class and location distribution:}
%In this work we use a softmax aggregation to summarize all scores intoe a single probability distribution
In our model, the last convolutional layer is a feature map $f$ with $H \times W=L$ locations and $C$ channels equivalent to the number of classes to classify. 
As shown in Fig.~\ref{fig:fig-0-convolution}(b), we can consider this feature map as a voting for the most likely position and class in an image. We can thus convert this feature map into a multinomial probability distribution over classes and position by applying a softmax on the two spatial dimensions and on the channels too. As we want each class and location to compete, we need to compute a single softmax on the three dimensions. 
% pure technical details. not necessary here.
%To implement in Pytorch, we first reshape the feature map into a single dimension, apply softmax on that dimension and then reshape it back to the original feature map shape.
Thus, instead of the common distribution over classes $p(c)$ as for classification, here we model the output of the CNN as a joint probability over classes and image locations: %\focus{you can use the notation ${f[c, l]}$ instead of ${f_{c,l}}$ to indicate the value in the row c and column l of the matrix f. }
\begin{equation}
    p(c,l) = \frac{\exp(f_{c,l})}{\sum_{c'=1,l'=1}^{C,L}  \exp(f_{c',l'})} \; .
    \label{equ:softmax}
\end{equation}
With this joint probability distribution we can obtain the class labels by marginalizing over locations: $p(c)=\sum_l p(c,l)$. This can be used to train our model for classification with standard cross-entropy loss. However, with the joint probability, we can also obtain the maximum a posteriori (MAP) of the best location $l^*$ and class $c^*$ for a given image: $c^*,l^* = \textrm{arg max}_{c,l} p(c,l)$. This is the information required to estimate the location and class of the object of interest. %Note that the model can automatically adapt the smoothness of the distribution based on the feature map scores. If for instance the object takes a small part of the image we would like a sharp distribution that is almost zero everywhere and close to the maximum at the right location. In contrast,   
%sharp or smooth distribution based on the 
This approach is simple and works well to find the center of the object. However, we are interested in yielding the bounding box of the object in the image. We can consider the bounding box of the object as proportional to the receptive field of the used feature map. However, this would lead to square bounding boxes at the same scale. To overcome the scale problem, in the next section we extend our approach to a multi-scale representation. %, however, the localization of the object is still poor because it is based on a fixed feature map

%The advantage of using softmax over other kind of aggregations is that it implicitly learn the most appropriate way to aggregate scores. Softmax can be parametrized by an inverse temperature $\gamma$ as follow:
%If $\gamma$ is close to 0 each score will have a similar probability 

\subsection{Multiscale search:}
For searching at multiple scales we use feature pyramids \cite{fpn-Lin-2017}, because it does not add much computational cost to the method and it works quite well on several problems.
With the feature pyramid, instead of considering a single feature map $f_{c,l}$, we use a representation composed by $S$ feature maps, each representing the image at a different scale. 
Thus, we can extend our joint distribution to also scales: $p(c,l,s)$ (see Fig.~\ref{fig:fig-0-convolution}(c)). Again, by marginalizing over locations and scales we can obtain $p(c)$ used for training, and by selecting the MAP, we can find the location $l^*$ and $s^*$ of the object of interest. %Although in principle a very similar extension to the previous 
Now, we can find objects at different scales and different locations. However, still, all objects will have the same aspect ratio. A possible solution would be to use convolutional filters of different sizes that will generate different receptive fields and therefore different bounding box shapes. However, this approach will increase the computational cost and will be able to provide only discrete object sizes (defined by the convolutional filters aspect ratio). In the next subsection we show how to learn a weakly supervised model that can adapt to any object size and aspect ratio.

% \begin{figure*}[tb]
%  \center
%  \includegraphics[height=4.5cm, width=0.8\linewidth]{images/multi_scale_architecture.png}
% \caption{Architecture of the proposed multi-scale convolutional STN for WSOL. Here we used the top two levels of the convnet(more levels can be used for better scale coverage). \focus{this figure will be removed.}}\label{conv_stn_model}
% \end{figure*}

% \begin{figure}
%  \center
%  \includegraphics[height=5.5cm, width=0.8\linewidth]{images/softmax_aggregation.png}
% \caption{softmax aggregation of the multi-scale conv STN. \focus{this figure will be redrawn.}}
% \label{softmax_aggregation}
% \end{figure}

% \begin{figure}[h!]
%   \centering
%   \includegraphics[width=1.\linewidth]{images/flatten} 
%   \caption{\texttt{Softmax} aggregation of the multi-scale CSTN.}
%   \label{softmax_aggregation}
% \end{figure}

\begin{figure*}
  \begin{center}
      \fbox{\includegraphics[height=5.5cm, width=0.7\linewidth]{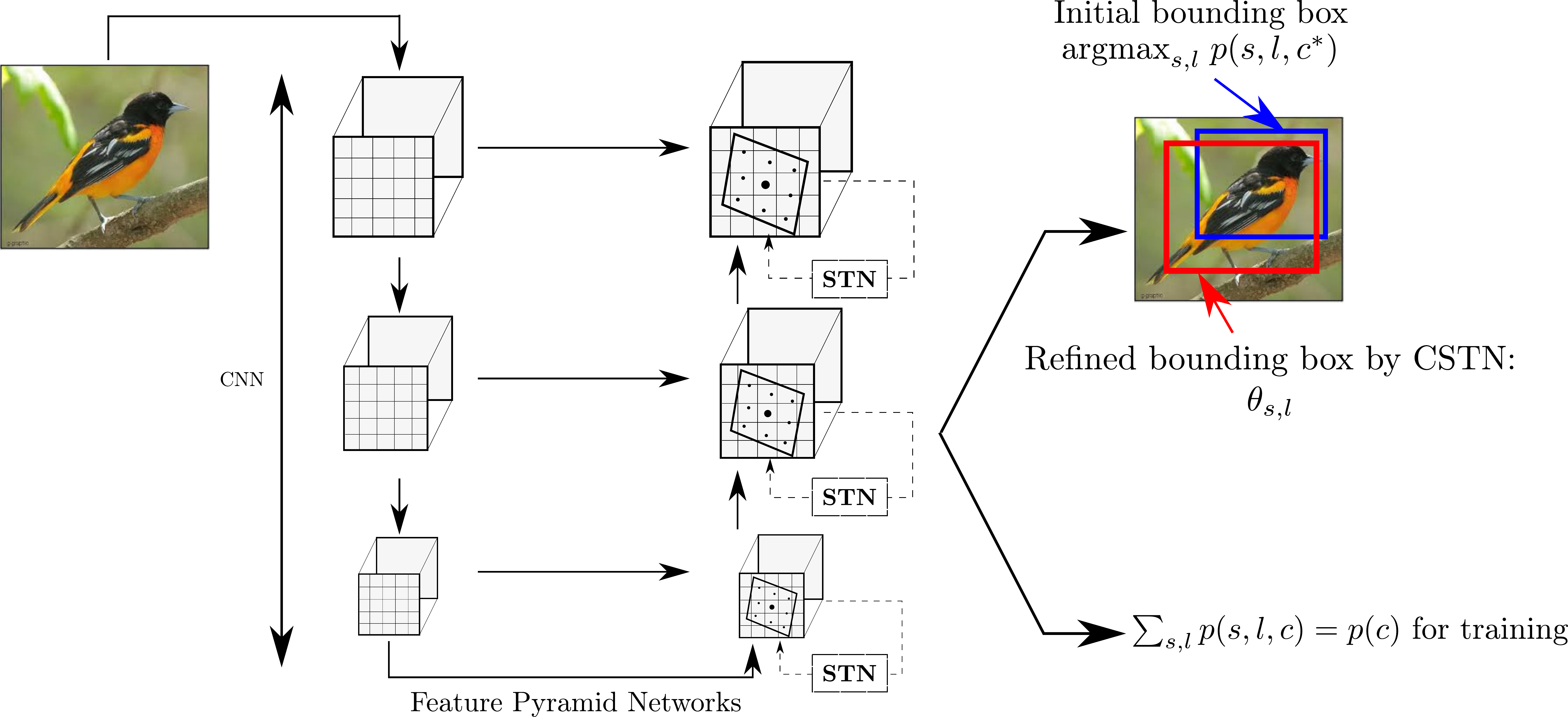}}
  \end{center}
  \caption{\textbf{Overall System.} This figure illustrates how all the components of the system are used in order to train the model in a weakly supervised manner as well as to perform inference.}
  \label{fig:system}
\end{figure*}

\subsection{Convolutional STN:}
While in fully supervised object detection most of the approaches regress a bounding box with the right object size, for weakly supervised models it is not possible because there is no ground truth to regress. In the original spatial transformer network (STN), a localization network is trained to find global image transformations that can better represent the data and therefore minimize the training loss. The authors of the original paper \cite{stn-Jaderberg2015} show that this is an approach for improving the classification performance by focusing on the object of interest and at the same time, being able to localize the object of interest without annotations, thus in a weakly supervised manner. 
However, we note that STN works well when the data is quite clean (e.g:, extended MNIST) and the sought transformations are relatively small. This is because the localization network of STN is trained with gradient descent, which is a local optimization. This means that when the transformation is too large or there is too much noise in the image, the local optimization will not be able to regress the correct transformation to localize the object and the training will fail. To overcome this problem, we propose to apply STN in a convolutional fashion. As shown in Fig.\ref{fig:fig-0-convolution}(d), for each feature map location we apply a localization network that reads the local features and generates a transformation based on those. As the STN is applied locally to each part of the image, the required transformation is smaller and it is more likely that the simple gradient based optimization used will work.
Thus, in this work, the last layer is now composed of two stages: i) estimation of the local transformations $\theta = loc(f)$, in which $loc$ is a convolutional localization network that for each feature map location $f_l$ returns a corresponding transformation $\theta_l$. 
ii) the final representation $f'$ is the results of a convolution in which the convolutional filters are now applied with the feature map transformations $\theta$: $f' = conv(f,\theta)$. 
The new layer is not much more expensive than a normal convolution because the additional computation is due only to the localization network. In contrast, being able to adapt the receptive field of the network to the local content of the image improves not only object localization but also the image classification. Even though powerful, in the experimental evaluation we note that the convolutional spatial transformer tends quite easily to overfit the training data. To avoid that in the next subsection we present two regularization techniques.

%However, by applying spatial transformer \cite{stn} in a convolutional way, we can regress the right size object boxes even without annotations.
\subsection{Regularization:}
%\end{comment}
Our multi-scale convolutional STN tends to focus on small regions. This is because during training, the selected bounding boxes shrink to the most discriminative part of an object while the classification performance improves. To address this, we added a regularization/penalty term to the classification loss which prevents the affine transformation $\theta_i$ from having large deviations from its reference location $\theta_{ref}$. This regularization term is,
\begin{equation}
L_{\theta}  = \sum_{s \in S} \sum_{i=1}^{h_s \times w_s}||\theta_{ref} - \theta_i||^2 \; .
\end{equation}
Here we choose $\theta_{ref} = \left[ \begin{matrix}
1 & 0 & 0 \\
0 & 1 & 0
\end{matrix} \right]$ corresponding to the identity transform. %Note that the $\theta_{ref}$ can be chosen to learn the transform with respect to different aspect ratios. Here our $\theta_{ref}$ corresponds to aspect ratio 1 for a $k \times k$ convolutional STN. Similarly for $\theta_{ref} = \left[ \begin{matrix}
%0.5 & 0 & 0 \\
%0 & 1 & 0
%\end{matrix} \right]$ the bounding box aspect ratio will be close to 0.5 and for $\theta_{ref} = \left[ \begin{matrix}
%1 & 0 & 0 \\
%0 & 0.5 & 0
%\end{matrix} \right]$, the aspect ratio will be close to 2. Learning the transform this way is similar to learning to regress with respect to anchor boxes in Faster R-CNN\cite{faster_rcnn}. Learning multiple $\theta_i$'s at each position of the convolution with respect to multiple aspect ratios is ideal, but here for simplicity here we used only identity transform for aspect ratio 1.

The multi-scale search has also a bias towards localizing large objects from the lower levels of a feature pyramid. It is due to the fact that, in many cases, object parts are more discriminative than the entire object; lower level layers will get strong activation for object parts of the large objects. %\sbx{Needs better justification as raised by r-cvpr. take some explanations from cvpr rebuttal to better support this sentence.}. %Often the classifier at this level tends to prefer  regions of the large object that triggers this issue. 
In order to make the higher levels compete for localizing large objects, we enforce the difference between the maximum activation of the two levels to be zero or negative, such that the higher feature map will be more likely to be selected. This can be applied on any two scale-adjacent feature maps $s_1$ and $s_2$,
\begin{multline}
L_{scale}(x)  = \max \bigg(0, \max_l p(s=s_1, l,c=c^*|x) - \\
 \max_l(p(s=s_2, l, c=c^*|x) \bigg)
\end{multline}
Notice that for small objects which gets localized from the lower level, this does not induce any penalty. Though the competitiveness among the levels can be ensured in many ways, this simple regularization term has given satisfactory results in our experiments. %benefit of mutli-scale search in our study.

With these regularization terms, the final loss function optimized by our model is:
\begin{equation}
L(x,y) = L_{cls}(x, y) + \lambda L_{\theta} + \alpha L_{scale}(x)
\end{equation}
where $L_{cls}(x, y)$ is the multi-class cross-entropy loss, $\alpha$ and $\lambda$ are hyper-parameters to specify the strength of the STL and multi-scale regularizations. 

\subsection{Complete System}
Fig.\ref{fig:system} summarizes our complete system. Given an image, a feature pyramid network builds semantic representations of the image at different scales. On all the scales, a CSTN is applied so that for each location and scale a localization bounding box is estimated. Finally the scores of the STN are converted in a joint probability $p(c,l,s)$ over classes, locations and scales. This can be converted to $p(c)$ by marginalizing over scales and locations to obtain the class probabilities needed to train the model in a weakly supervised manner. During training the proposed regularizations are also used. The joint probability is used at test time to localize the object by a MAP inference.

% !TEX root=main.tex

\section{Experiments}
\label{sec:experiments}

\subsection{Experimental setup:}

\noindent \textbf{Datasets and evaluation metric}.
We evaluated our multiscale convolutional STN model on CUB-200-2011 dataset \cite{cub_dataset-Welinder-2010} and ILSVRC 2012 \cite{imagenet-Russakovsky-2015} localization dataset. CUB-200-2011 contains 11,788 images of 200 bird species with 5,994 images for training and 5,794 for testing. %Since this dataset is fine-grained classification problem, discriminating birds categories is challenging and it is a good benchmark for our model. 
ILSVRC 2012 dataset contains 1.28M training images and 50,000 validation images. There are 1000 categories of objects. 
For both datasets we evaluate the performance in terms of classification and localization accuracy. An image is said to be correctly localized if the predicted class matches the true class and the predicted bounding box has 50\% overlap with the ground-truth. The localization accuracy is denoted as Top-1 Loc in the results. For explicitly measuring the localization performance(regardless of the classification accuracy), another metric called GT-Known Loc is used where the GT image label is provided. In that case, a localization is deemed as correct if the 50\% overlap criteria is satisfied. Unlike CAM, our method can provide multiple bounding boxes per image. But in Top-1 Loc we are only using the box with the highest score. When this  top scoring box is centered at the object, we get the best bounding box. But this is not always the case with CSTN, especially when the top scoring box is focusing on the discriminative object regions. There could be other boxes, for which the score is very close to the top box but they overlap well with the ground-truth box. So we also considered a metric which we call the Top-5 box localization where we check if one among the top five boxes with high scores has 50\% overlap with the object. Top-5 box localization gives interesting results regarding the localization ability of our method in contrast to the CAM. We measured GT-Known Top-5 box Loc in this comparison.

\noindent \textbf{Implementation details}. 
We used ResNet101 \cite{resnet-He-2016} as the backbone network which is pre-trained on ImageNet \cite{imagenet-Russakovsky-2015}. We removed the last average pooling and fully connected layer and added an additional convolution(with $3 \times 3$ filter size and padding 1) and batch norm \cite{batch_norm-Ioffe-2015} layer. %The parameters of this layer are initialized randomly as in \cite{}. 
Feature pyramid is obtained from this network with its last two levels as described in \cite{fpn-Lin-2017}. The input images are resized to $320 \times 320$ pixels. For data augmentation, we used horizontal flip with 50\% probability. Images are normalized with $\textrm{mean} = [0.485, 0.456, 0.406]$ and $\textrm{std} = [0.229, 0.224, 0.225]$ as in ImageNet training \cite{imagenet-Russakovsky-2015}. The model is trained on NVIDIA GTX 1080 GPU with 12GB memory.

%\focus{Due to space limitation, we provide as a supplementary material an ablation study of our method that includes: ..., ..., ...}
\subsection{Ablation Study:}
\begin{figure*}[h!]
\begin{center}
\fbox{\includegraphics[height=6.5cm, width=0.9\linewidth]{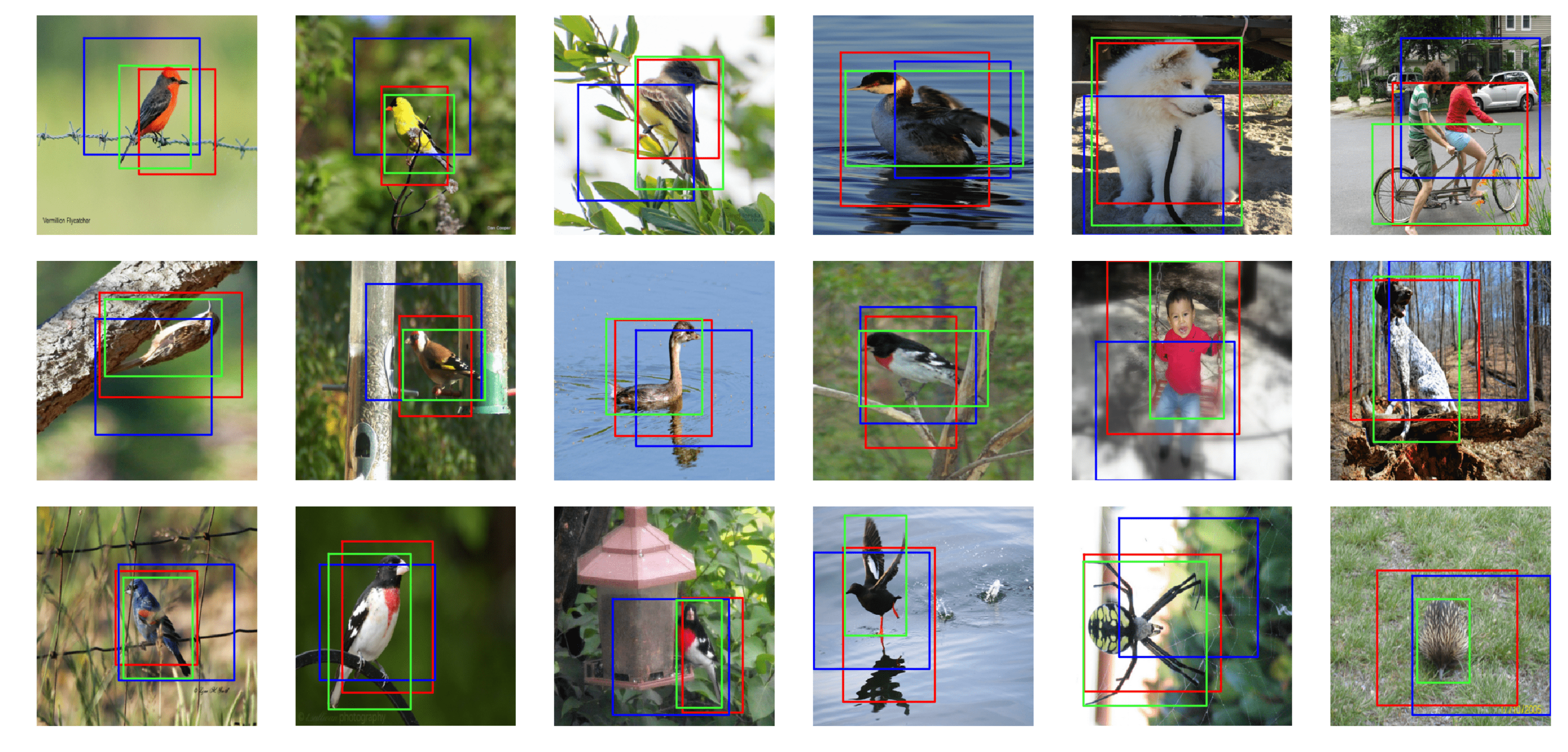}}
\end{center}
\caption{Demonstration of some transforms learned by CSTN on CUB-200-2011 and ILSVRC dataset. The last column shows some failed localization on the ILSVRC dataset. The non transformed box is shown in blue, transformed box in red and the ground-truth is green.}
\label{transform_adapt}
\end{figure*}

The ablation studies are conducted to assess the impact of spatial transform, multi-scale localization and the regularization on $\theta$. We used CUB-200-2011 dataset in our ablation experiments. To assess the importance of the spatial transform, we computed the localization accuracy when the default box is used for localization instead of the transformed output from STN. Note that this does not change the training procedure, since CSTN is still used the same way to learn the localization. At the implementation level, instead of using the transformed coordinates, we used the original coordinates to compute the localization performance. Table \ref{transform_impact} shows the result of this study on both datasets. It can be observed that the transform is improving the localization around 5-8\%. To see this impact visually, figure \ref{transform_adapt} shows some sample images where the transform is modifying the original receptive field box to improve the localization. It also highlights some failure cases where the transform is producing wrong localization.
\begin{table}[h!]
\caption{Impact of transform on the localization performance. For both datasets the CSTN is fundamental to obtain good performance.}
\label{transform_impact}
\begin{center}
    \begin{tabular}{|l|c|c|}
        \hline
        \textbf{Dataset} & \multicolumn{2}{c|}{\textbf{Top-1 Loc}}  \\
        \cline{2-3}
          & without transform & with transform\\
        \hline
        CUB-200-2011 & 40.64 & 49.03 \\
        ImageNet & 36.69 & 42.38\\
        \hline
    \end{tabular}
\end{center}
\end{table}

To further study whether the CSTN is learning a good representation for localization we compared the localization performance with and without CSTN. For the case without CSTN, we used the same architecture and classification head, the only difference is that no transform is learned in this settings, i.e, instead of CSTN, a normal convolution is used. The classification head is now classifying the fixed sampling space of the convolution. Table \ref{no_transform}  shows the result of this study on CUB-200-2011 dataset. It can be observed that without CSTN, the localization performance has reduced drastically. The classification performance also goes low but the impact is less. This means that the CSTN not only learn to better localize an object in the image, but it also learn a better representation of the object that produces an improved classification.

\begin{table}[h!]
\caption{Localizing with and without convolutional STN on CUB-200-2011 dataset. It can be observed that the CSTN is very effective in learning a good representation for localization. It improves the localization by 26.79\%.}
\label{no_transform}
\begin{center}
    \begin{tabular}{|l|c|c|}
        \hline
        \textbf{Type} & \textbf{Top-1 Class} & \textbf{Top-1 Loc}  \\
        \hline
        Without conv STN & 77.40 & 21.64 \\
        With conv STN & 78.46 & 49.03\\
        \hline
    \end{tabular}
\end{center}
\end{table}

\begin{figure}[!t]
 \centering
    \subfloat[Localization from level 4 (Top-1 Loc is 37.78\%).]{\includegraphics[height=3.8cm, width=\linewidth]{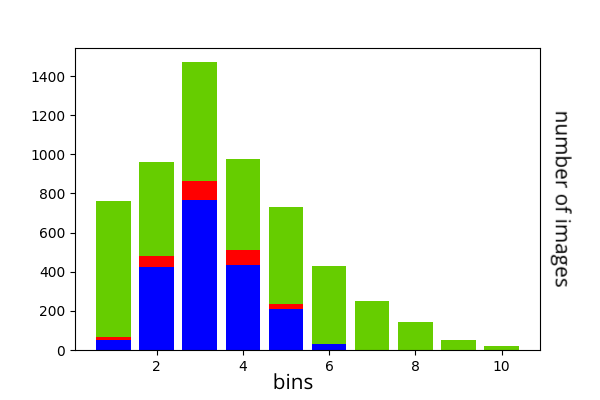}} \newline
    \subfloat[Localization from level 5 (Top-1 Loc is 42.91\%).]{\includegraphics[height=3.8cm, width=\linewidth]{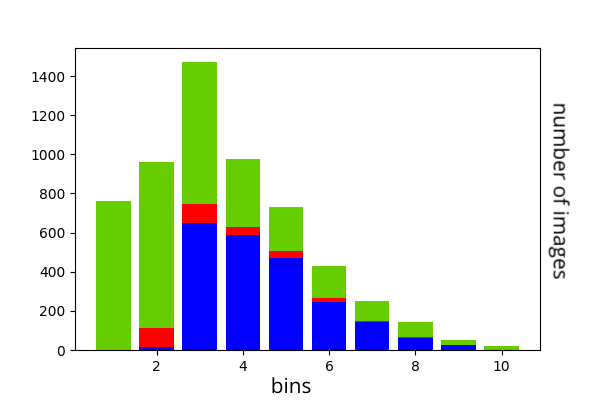}} \newline
    \subfloat[Localization with the multi-scale model combining levels 4 and 5 (Top-1 Loc is 48.43\%).]{\includegraphics[height=3.8cm, width=\linewidth]{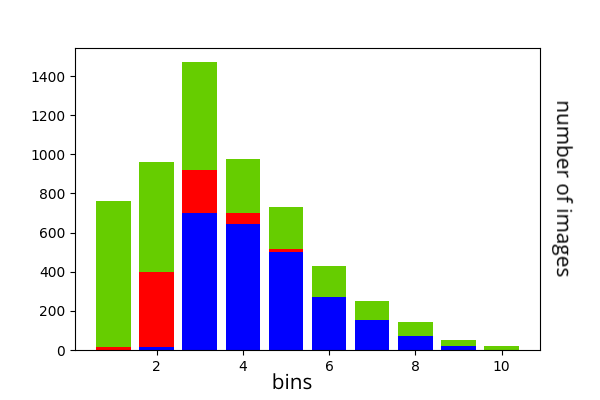}}
\caption{Impact of multi-scale localization. Localization from each level is compared with the multi-scale model which combines all levels. The histogram is created by dividing the area of all bounding boxes into 10 equal bins. Green bars shows the number of images in each bin, red bar shows number of images that are correctly localized by CSTN in that bin and the blue bars shows the number of images correctly localized without the bounding box transformation, as reported in Table I.}
\label{multi_scale_ablation}
\end{figure}

The multi-scale localization is another important component in our model. To assess the importance of this, we conduct ablation experiments with localization from two level of the feature pyramid independently and compare it with the model where these levels are combined. Figure~\ref{multi_scale_ablation} shows the results from this study. Here a histogram is created by dividing the area of the bounding box into 10 bins of equal size. The histogram shows in green the total number of samples at each resolution and in blue and red the percentage of images that are correctly localized in each bin for the model without and with bounding box transformations. From figure \ref{multi_scale_ablation}(a) and \ref{multi_scale_ablation}(b) we see that different levels are specialized on different object sizes. With the multi-scale model (Figure \ref{multi_scale_ablation}(c)) we balance the localization between the two levels and improve the localization accuracy. Notice also that the effect of the bounding box transformation become stronger when using a multi-scale model. This is in line with out hypothesis that the STN performs a local optimization and for improved performance, the transformations should be relatively small from a reference size. This can be compared to learning the transforms with respect to anchor boxes in the fully supervised object detectors\cite{faster_rcnn-Ren-2015}.

 8

\begin{figure}
 \center
 \includegraphics[height=4.7cm, width=0.9\linewidth]{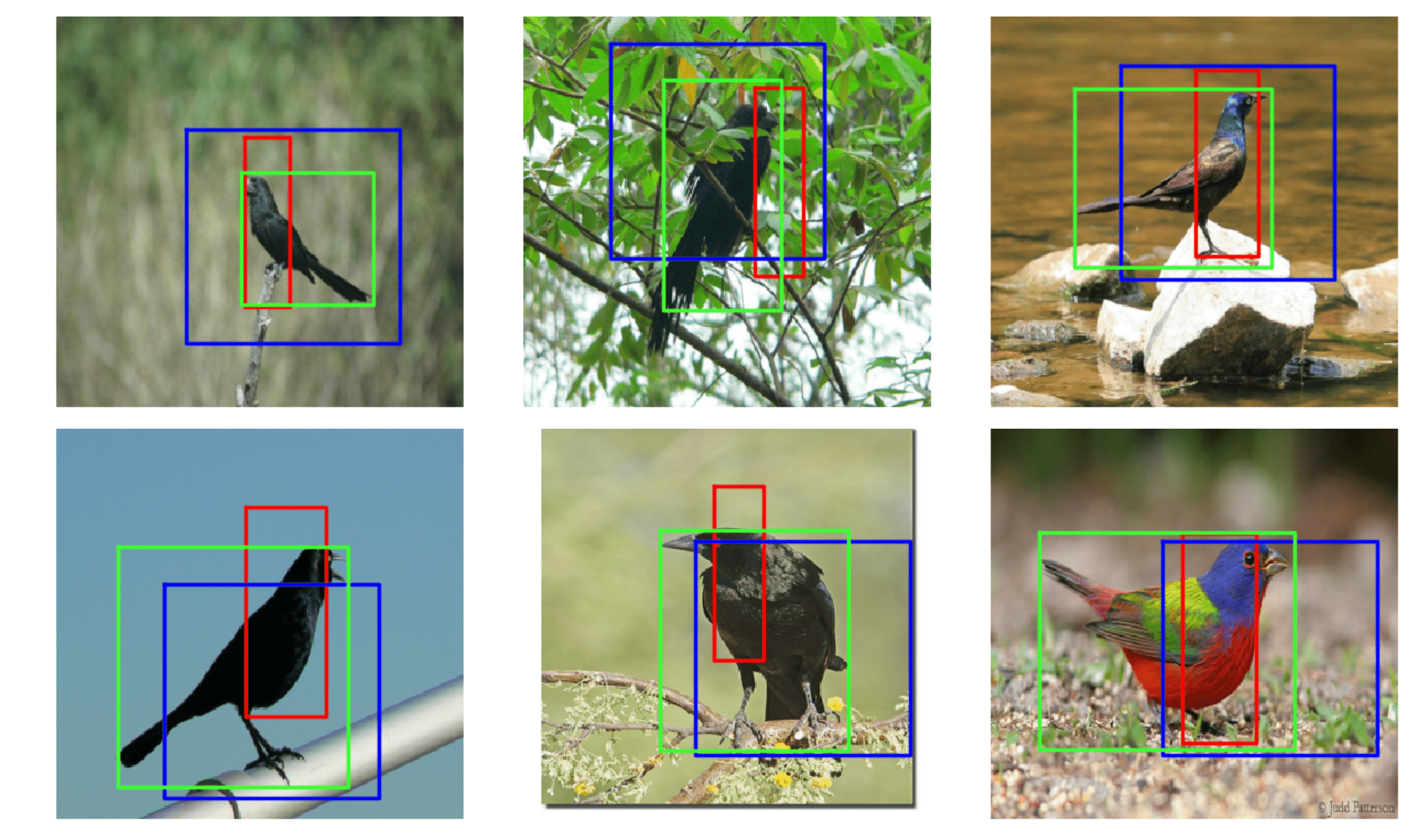}
\caption{Transforms learned without using the regularization on $\theta$. The receptive field box is shown in blue, transformed box in red and the ground-truth is green. It can be observed that, the boxes learned are not from the distribution of possible object bounding boxes.}
\label{no_theta_reg}
\end{figure}

Another key component of our method is the regularization on $\theta$. We observed that without this regularization, the learned transformations are not from the distribution of possible object bounding boxes. The transformations tend to overfit and shrink to discriminative image parts resulting in poor localization. Figure \ref{no_theta_reg} show samples of bounding boxes learned without using regularization on $\theta$. To obtain a good localization, tuning the hyperparameter $\lambda$ is critical. Table \ref{theta_reg_tuning} shows the performance in classification and localization for different values of $\lambda$. As expected, while the model classification is barely affected, localization is highly affected by this parameter. 
For the regularization on the scales, we found that $\alpha$ can vary in a range of values without affecting too much the localization results. Thus we did not include a study on that.

\begin{table}[h]
\caption{Impact of $\lambda$ on classification and localization accuracy. For a high value of $\lambda$ the localization accuracy tends to the one obtained without STN. For no regularization, the transformations become too strong and focus on small parts of the object thus producing a very poor localization score. }
\label{theta_reg_tuning}
\centering
    \begin{tabular}{|l | c | c|}
        \hline
        $\lambda$ &  \textbf{Top-1 Loc} & \textbf{Top-1 Class} \\
        \hline
        0.01 & 27.39 & 78.98\\
        0.001 & 30.88 & 78.63\\
        0.0001 & 49.03 & 78.46\\
        0.00001 & 45.52 & 78.25\\
        0 & 5.13 & 77.32 \\
        \hline
    \end{tabular}
\end{table}

\subsection{Comparison with state-of-the-art methods:}
We compare the localization of the CSTN with state-of-the-art solutions for WSOL. Results are summarized in table \ref{cub_sota_compare} and table \ref{imagenet_sota_compare} for CUB-200-2011 and ILSVRC 2012 respectively. 

On the CUB-200-2011 dataset, CSTN performs better than all the CAM based methods. except the ADL \cite{adl-Choe-2019}. 
In this dataset, the scale of objects are distributed unevenly, i.e, many objects are of nearly the same size, extreme variations in the size are very less (not too many small and large objects). As shown in the ablation study, different levels of the CSTN specializes on different scales, therefore, we can get the best of the localization from this model by focusing more on the crowded scales (where there are many objects). The hyper-parameter $\alpha$ is not very sensitive to the Top-1 Loc, so it can be tuned fairly easily. The difference in performance with ADL is mostly due to the wrong location selection as the recall is still close to 99\%(so the CSTN is able to produce transformations that match the object sizes). The GT-Known Top-5 Loc is around \textbf{2.5\%} higher than the GT-Known Loc of the ADL. This also reinforces our claim that the CSTN is learning better localization. Compared to all the CAM based methods including the state-of-the-art ADL\cite{adl-Choe-2019}, CSTN has some clear advantages. These methods need rigorous tuning of their hyperparameters to obtain good localization. The hyperparameters of CSTN are not very sensitive to the localization accuracy. The $\alpha$ regularization term can be avoided if we can find a suitable heuristic that tells whether the object is small or large. Thus small object can be localized from lower level and the large one from higher level. Then it works similar to the level selection in multi-scale fully supervised detectors based on the area of the ground truth box. The $\theta$ regularization is also fairly easy to tune as shown in the ablation experiments.
Recent studies observed that WSOL algorithms which improve the localization based on erase and learn strategy\cite{acol-Zhang-2018,adl-Choe-2019, hide_and_seek-Singh-2017} are very sensitive to their hyperparameters\cite{eval_wsol-Choe-2020}.
%ADL \cite{adl-Choe-2019} is proposing a regularization strategy to identify more discriminative regions from a classification network. Since it is generic, it can be applied in our method also to improve the localization further. Here we want to test the  localization ability of the CSTN, so we used the basic architecture in our experiments. 

\begin{figure}[!t]
\begin{center}
\fbox{\includegraphics[height=4cm, width=0.8\linewidth]{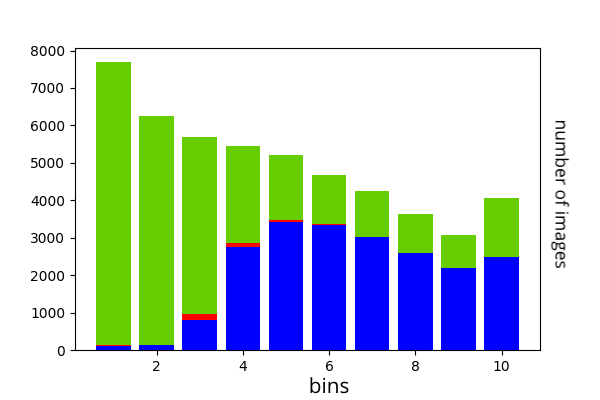}}
\end{center}
\caption{Histogram of localization on ImageNet validation set. The histogram is created by uniformly dividing the range of the area of objects in the validation set. It can be observed that the scale of objects are not uniform.}
\label{imagenet_multiscale}
\end{figure}

On the ILSVRC dataset, CSTN is outperformed by many of the CAM based method. This is probably due to the sensitivity to the scale. The number of objects in different scales are nearly uniformly distributed in this dataset. So the multi-scale localization should specialize on each scale equally well in this case. This can be better explained with the histogram of localization on ImageNet shown in figure \ref{imagenet_multiscale}. As we can see, it favors the localization towards large objects in this case. As a result, it fails to localize most of the smaller objects. The GT-Known Top-5 Loc in this case is comparable to the GT-Known Loc of the state-of-the-art methods including ADL and SPG.

\begin{table}
\caption{Performance comparison on the CUB-200-2011
test set. Convolutional STN performs better than all other methods, except ADL. The Top-1 class is left blank for some methods, because it is not reported in the original paper.}
\label{cub_sota_compare}
\centering
    \begin{tabular}{|l|c|c|c|}
        \hline
        \textbf{Method} & \textbf{Top-1 Loc} & \textbf{GT-Known Loc} & \textbf{Top-1 Class} \\
        \hline \hline
        CAM \cite{cam-Zhou-2016} & 41.00 & 71.13 & - \\
        %\cite{cam-Zhou-2016} & 42.72 & 80.65 \\
        HaS \cite{adl-Choe-2019,hide_and_seek-Singh-2017} & 44.67 & 73.32 & 76.64 \\
        ACoL \cite{acol-Zhang-2018} & 45.92 & 75.30 & 71.90 \\
        SPG \cite{spg-Zhang-2018} & 46.64 & 74.11 & - \\
        ADL \cite{adl-Choe-2019} & \textbf{62.29} & 78.62 & 80.34\\
        CSTN & 49.03 & 76.06 & 78.46\\
        \hline
        \textit{CSTN Top-5 box} & - & 81.14 & - \\
        \hline
    \end{tabular}
\end{table}

\begin{table}
\caption{Performance comparison on the ILSVRC validation set. The Top-1 Loc is competitive but due to the sensitiveness to scale, convolutional STN miss to localize small objects. The sensitivity of the CAM to scale is less, so this can be the reason for the difference in Top-1 Loc.} %\sbx{Comparison to \cite{adl-Choe-2019} is missing. suggested by cvpr-reviewers.}}
\label{imagenet_sota_compare}
\centering
    \begin{tabular}{|l|c|c|c|}
        \hline
        \textbf{Method} & \textbf{Top-1 Loc} & \textbf{GT-Known Loc} & \textbf{Top-1 Class} \\
        \hline \hline
        CAM \cite{cam-Zhou-2016} & 42.80 & 61.10 & 66.60 \\
        %CAM ResNet50-SE \cite{cam-Zhou-2016} & 46.19 & 76.56 \\
        %HaS \cite{adl-Choe-2019,hide_and_seek-Singh-2017} & 41.87 & 63.12 & 67.48 \\
        HaS \cite{hide_and_seek-Singh-2017} & 45.21 & 63.12 & 70.70 \\
        ACoL \cite{acol-Zhang-2018} & 45.83 & 62.73 & 67.50 \\
        SPG \cite{spg-Zhang-2018} & \textbf{48.60} & 64.24 & - \\
        ADL \cite{adl-Choe-2019} & 48.43 & 63.72 & 75.85\\
        CSTN & 42.38 & 60.48 & 69.48\\
        \hline
        \textit{CSTN Top-5 box} & - & 63.45 & - \\
        \hline
    \end{tabular}
\end{table}

\begin{figure}
\begin{center}
\fbox{ \includegraphics[height=4cm, width=0.9\linewidth]{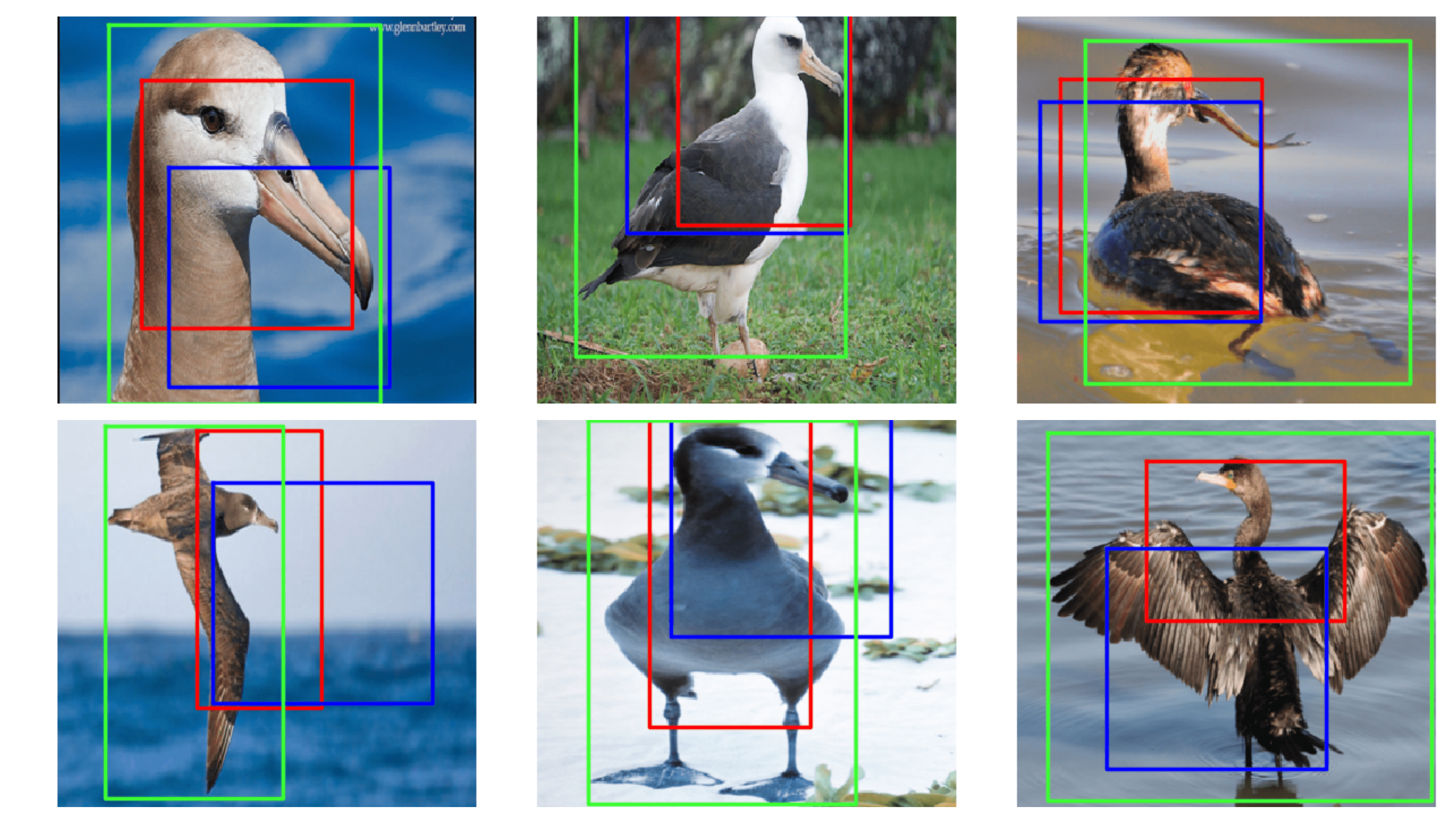}}
\end{center}
\caption{Localizing large objects using the wrong scale. The STN fails to learn large transforms for this case to give an accurate localization The receptive field box is shown in blue, transformed box in red and the ground-truth is green.}
\label{bad_scale}
\end{figure}

We believe that improving the multi-scale localization component of our method can close this performance gap compared to the state-of-the-art CAM based WSOL. If we try to localize an object with the wrong scale (i.e, from the wrong level of the feature pyramid), it will end-up in getting stuck at some discriminative object region. Figure \ref{bad_scale} shows some failure cases of this when localizing large objects using CSTN. Since the end goal of STN is still to get a good classification, it will not try to localize the integral object. The softmax aggregation strategy is a simple and straightforward expansion to introduce the multi-scale capability. Having better methods to select the matching scale can bring the benefit of CSTN to all such multi-scale improvements. Moreover, improving the box selection strategy can also give better Top-1 Loc, since our GT-Known Top-5 Loc is always good.

% !TEX root=main.tex

\section{Conclusion}
\label{sec:conclusion}

In this work, we have introduced a novel method for weakly supervised object localization. Different from the dominant paradigm of Class Activation Maps, we show that the use of a  convolutional spatial transformer can lead to competitive performance in localization. Compared to the activation map based methods, convolutional spatial transformer is less sensitive to their hyperparameters for weakly supervised localization. This component can be plugged into any convolutional network giving an end-to-end weakly supervised localization module. The learning of the convolutional STN is fairly easy and it adds few additional convolutional layers to the standard CNN. Our Convolutional STN with multi-scale localization gives competitive results on the benchmarked datasets. Empirical study reveals that the localization with convolutional STN is sensitive to the object scale and we have proposed two regularization strategies to deal with those issues. Future work is about extending the method to weakly supervised object detection. 
%We have used a simple multi-scale localization here with softmax integration. For future work, we will device better solutions to handle the scale issue. 


%\section*{Acknowledgment}



%\bibliographystyle{IEEEtran}
%\begin{thebibliography}{1}

%H.~Kopka and P.~W. Daly, \emph{A Guide to \LaTeX}, %3rd~ed.\hskip 1em plus

%\end{thebibliography}
\bibliographystyle{./IEEEtran}
\bibliography{./IEEEabrv,./IEEEexample}

\end{document}